\documentclass[letterpaper, 10 pt, conference]{ieeeconf}  

\IEEEoverridecommandlockouts                              

\usepackage{caption}
\usepackage{amsmath,amsfonts}
\usepackage{upgreek}
\usepackage{algorithm}
\usepackage{algpseudocode}
\usepackage{array}
\usepackage{amssymb}
\usepackage{textcomp}
\usepackage{stfloats}
\usepackage{url}
\usepackage{verbatim}
\usepackage{graphicx}
\usepackage{cite}
\usepackage{subcaption}
\usepackage{soul}
\usepackage{color, xcolor}
\usepackage{multirow}
\usepackage{makecell} 
\usepackage{stfloats}
 \usepackage{booktabs}
 \usepackage{hyperref}
\soulregister{\cite}7
\soulregister{\citep}7
\soulregister{\citet}7 
\soulregister{\ref}7
\soulregister{\pageref}7

\title{\LARGE \bf
Automated 3D-GS Registration and Fusion via Skeleton Alignment and Gaussian-Adaptive Features
}

\author{Shiyang Liu$^{1}$, Dianyi Yang$^{1}$, Yu Gao$^{1}$, Bohan Ren$^{1}$, Yi Yang*$^{1}$, Mengyin Fu$^{1}$
\thanks{This work was partly supported by National Natural Science Foundation of China (Grant No. NSFC 62233002) and National Key R\&D Program of China (2022YFC2603600). (*Corresponding Author: Y. Yang, yang\_yi@bit.edu.cn)}
\thanks{$^{1}$School of Automation, Beijing Institute of Technology, Beijing, China}%
}

\begin{document}

\maketitle
\thispagestyle{empty}
\pagestyle{empty}

\begin{abstract}
In recent years, 3D Gaussian Splatting (3D-GS)-based scene representation demonstrates significant potential in real-time rendering and training efficiency. However, most existing methods primarily focus on single-map reconstruction, while the registration and fusion of multiple 3D-GS sub-maps remain underexplored. Existing methods typically rely on manual intervention to select a reference sub-map as a template and use point cloud matching for registration. Moreover, hard-threshold filtering of 3D-GS primitives often degrades rendering quality after fusion. In this paper, we present a novel approach for automated 3D-GS sub-map alignment and fusion, eliminating the need for manual intervention while enhancing registration accuracy and fusion quality. First, we extract geometric skeletons across multiple scenes and leverage ellipsoid-aware convolution to capture 3D-GS attributes, facilitating robust scene registration. Second, we introduce a multi-factor Gaussian fusion strategy to mitigate the scene element loss caused by rigid thresholding. Experiments on the ScanNet-GSReg and our Coord datasets demonstrate the effectiveness of the proposed method in registration and fusion. For registration, it achieves a 41.9\% reduction in RRE on complex scenes, ensuring more precise pose estimation. For fusion, it improves PSNR by 10.11 dB, highlighting superior structural preservation. These results confirm its ability to enhance scene alignment and reconstruction fidelity, ensuring more consistent and accurate 3D scene representation for robotic perception and autonomous navigation.
\end{abstract}

\section{INTRODUCTION}

Multi-view collaborative mapping holds significant application value in unmanned systems \cite{lin2013cross}. With the advancement of 3D reconstruction, map representations in such systems have evolved substantially. Beyond conventional voxels, meshes, and point clouds, emerging paradigms like neural radiance fields (NeRF) \cite{mildenhall2021nerf, barron2021mip, gao2023mc} and 3D Gaussian Splatting (3D-GS)\cite{kerbl20233d, wu20244d, qin2024langsplat} have redefined environmental modeling. Among these, 3D-GS emerges as a promising solution for high-fidelity mapping by explicitly modeling radiative properties while addressing critical requirements in unmanned systems – striking a balance between real-time rendering and visual fidelity \cite{ zheng2024gaussiangrasper, wang2024reinforcement}. Compared to traditional discrete representations, 3D-GS achieves superior rendering quality while maintaining adaptability for multi-sensor fusion \cite{hong2024liv}, making it well-suited for applications demanding both precision and operational efficiency in complex environments.

\begin{figure}[t]
\centering
\includegraphics[width=\linewidth]{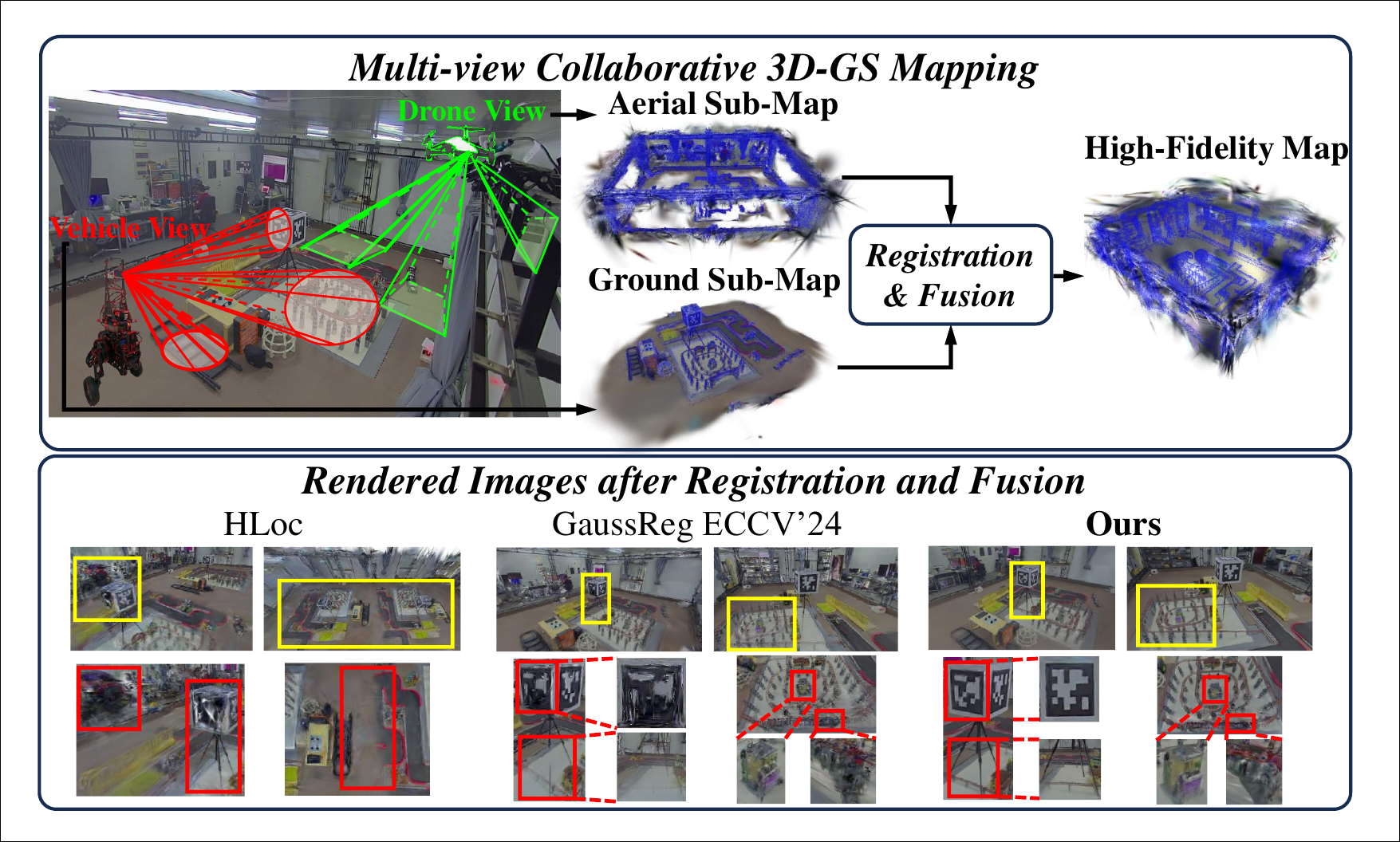} 
\captionsetup{font=small}
\caption{\textbf{Illustration of our task.} A vehicle (Diablo) and a drone (Tello) independently capture data and reconstruct separate 3D-GS sub-map. The drone view offers broad structural coverage, while the vehicle view captures fine details. A fusion strategy integrates both, producing a unified 3D-GS model with global completeness and high fidelity.}
\label{fig:story_pipeline}
\end{figure}

However, most existing 3D-GS-based mapping approaches focus on global or incremental reconstruction within a single map \cite{matsuki2024gaussian, yan2024gs}, with limited attention to the registration and fusion of multiple sub-maps. As illustrated in Fig.\ref{fig:story_pipeline}, when an unmanned system constructs two 3D-GS sub-maps from intersecting ground and aerial viewpoints, aligning and integrating them into a coherent and photorealistic 3D representation remains a challenging task. Unlike point cloud, 3D-GS registration requires jointly estimating spatial transformations while accounting for Gaussian-specific attributes such as anisotropic covariance, opacity, and spherical harmonics, which influence both geometric alignment and radiance consistency. Additionally, fusion involves optimizing Gaussian selection and blending strategies to avoid structural distortions and radiometric inconsistencies, ensuring seamless integration while preserving both global structure and fine-grained details for high-fidelity rendering.

A feasible solution has been to manually select the sub-map with relatively better geometric features as the foundational skeleton, standardize the coordinate system, and enforce registration by matching the other sub-map to it \cite{chang2024gaussreg, yuan2024photoreg, zhu2024loopsplat}. Subsequently, a spatial threshold is applied to determine which 3D-GS primitives should be retained from each sub-map, achieving scene fusion. Nevertheless, these approach faces the following challenges:

\textbf{1) Manually assisted geometric registration}: Manually determining which sub-map has better geometric structure can be brittle, especially when both exhibit similar rendering quality. Moreover, most methods now used registration methods mainly employ point cloud-based transformation approaches \cite{zhu2020leveraging, li2023learning, hou2024pos, liu2023tie, li2021point, yuan2023egst, qin2023geotransformer, poiesi2022learning, yu2023peal}, neglecting 3D-GS attributes including opacity, anisotropic covariance, and spherical harmonics, thereby limiting their capacity to utilize 3D-GS properties.

\textbf{2) Hardcore threshold truncation fusion}: After registration, applying hard distance threshold filtering to 3D-GS primitives may degrade rendering quality, potentially leading to issues such as hollow objects, missing parts, or incomplete scene representation \cite{chang2024gaussreg, shorinwa2025siren}.

To address the above issues, we propose an automated structure-aware framework that eliminates manual intervention and enhances scene registration and fusion. First, we introduce a geometry skeleton extraction and registration strategy, ensuring global structural consistency through automated merging and refinement. Second, to overcome limitations in feature extraction due to the anisotropic nature of 3D Gaussians, a Gaussian-adaptive convolution method is developed, leveraging Mahalanobis distance-based neighborhood selection and ellipsoid-aware kernels to enhance local feature representation \cite{mahalanobis2018generalized}. Finally, to preserve structural integrity and detail accuracy, we introduce a score-based Gaussian primitive selection strategy that balances structural consistency, local detail fidelity, and spatial coherence. In conclusion, our research contributions are as follows:

1) A skeleton-based initialization and Gaussian-adaptive feature extraction framework, integrating anisotropic neighborhood selection and ellipsoid-aware convolution to enhance feature distinctiveness, enabling efficient preprocessing and robust registration of sub-maps.  

2) An optimized Gaussian fusion strategy, incorporating multi-factor scoring to refine Gaussian selection, ensuring structural preservation and seamless integration.  

3) Extensive evaluations on ScanNet-GSReg dataset demonstrate competitive registration accuracy and reconstruction fidelity, validating the effectiveness of the proposed fusion strategy in preserving fine-grained details. 

\section{RELATED WORK}

\subsection{3D Scene Representation and Registration Methods}

Traditional 3D scene representations, including point clouds, voxel grids, and meshes, are widely used in robotics but struggle with sparse data, computational complexity, and noise propagation \cite{lai2025structural, rodrigues2013structured}. Classical registration methods, such as ICP \cite{li2020evaluation, liu2019point, vizzo2023kiss} and RANSAC \cite{cheng2023sampling, chung2024centralized}, rely on rigid transformations and geometric consistency, making them less effective in handling large viewpoint variations and structural discrepancies between independently reconstructed scenes. Learning-based approaches enhance robustness via deep feature matching \cite{lu2019deepvcp, lu2021hregnet} and semantic-aware networks \cite{wu2025evolutionary} but remain constrained by viewpoint sensitivity. Recent advancements in NeRF have introduced novel registration paradigms, shifting from purely geometric to radiance-based representations. NeRF2NeRF \cite{goli2023nerf2nerf} employs keypoint supervision for multi-modal alignment at the cost of scalability, while DReg-NeRF automates geometric disentanglement but struggles with large-scale adaptability \cite{chen2023dreg}. NeRFuser \cite{fang2023nerfuser} enhances multi-view consistency but inherits volumetric rendering overhead. Despite NeRF’s advances, its implicit nature obscures explicit geometric relationships, limiting robotic applications. 3D-GS mitigates this by combining geometric manipulability with differentiable photometric optimization, achieving faster registration and high-fidelity rendering crucial for robotic perception.

\subsection{3D-GS-Based Reconstruction and Registration }

Integrating aerial and ground-level data for 3D reconstruction presents challenges in resolution, scale consistency, and occlusions. Traditional methods, including multi-view stereo (MVS) \cite{gao2019ground}, photogrammetry \cite{qin20203d}, and mesh-based techniques \cite{zhu2020leveraging}, often suffer from misalignment due to significant viewpoint variations. 3D-GS-based approaches address these limitations by leveraging Gaussian representations for more flexible and robust fusion. UC-GS \cite{zhang2024drone} enhances scene fidelity by integrating aerial and vehicle views, while DRAGON \cite{ham2024dragon} iteratively aligns 3D Gaussians but relies on predefined poses, limiting adaptability. Pose-free solutions such as SelfGaussian \cite{kang2024selfsplat} incorporate 3D priors for registration, whereas Adaptive GS Fusion \cite{zhang2025crossview} introduces automated alignment strategies. However, existing approaches depend on manual initialization or strong scene priors, restricting their generalization. The proposed framework improves 3D-GS-based registration and fusion by introducing structure-aware skeleton alignment, Gaussian-adaptive feature extraction, and multi-factor fusion optimization, ensuring both global structural consistency and local detail preservation. These advancements enable seamless integration of independently reconstructed aerial and ground-based models, improving the accuracy and robustness of large-scale 3D scene representation.

\section{METHOD}

\begin{figure*}[htbp]
\centering
\includegraphics[width=0.67\linewidth]{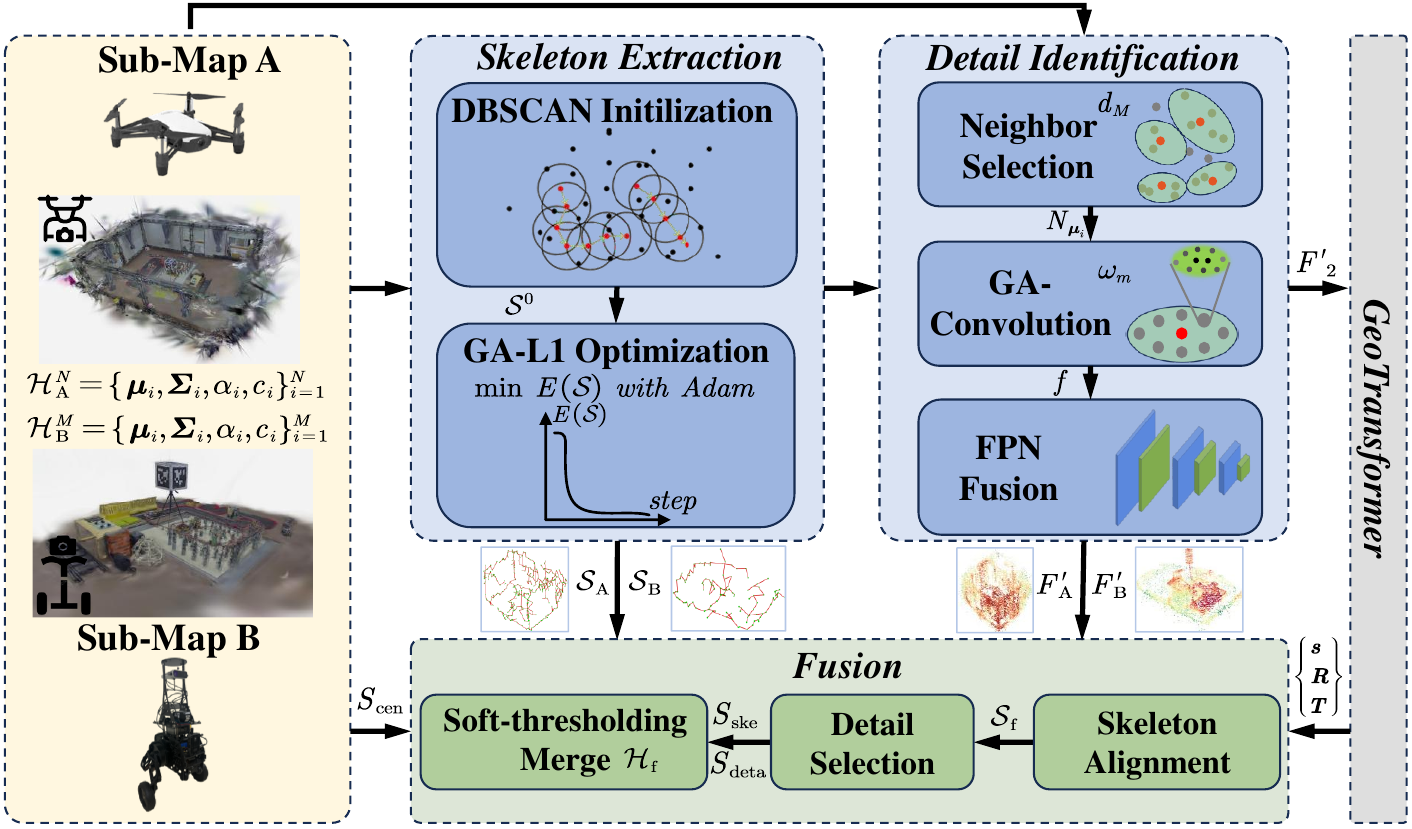} 
\captionsetup{font=small}
\caption{Overview of the proposed method. Given two sub-maps collected from different perspectives, we extract skeletons and local details from sub-maps, register them using GA-KPConv with GeoTransformer \cite{qin2023geotransformer}, which takes the first two layers’ output features from the sub-maps as input and estimates the transformation, and fuse Gaussians through a soft-thresholding strategy based on skeletons, local details, and scene-center distances.}
\label{fig:method_pipeline}
\end{figure*}

This section presents the proposed framework for 3D-GS registration and fusion, as illustrated in Fig.~\ref{fig:method_pipeline}. First, a Gaussian-skeleton distance is introduced to initialize and refine structural alignment. Then, Gaussian-Adaptive (GA)-KPConv enhances feature extraction with anisotropic neighborhood adaptation. Finally, a structure-aware fusion strategy integrates registered sub-maps while preserving geometric consistency and fine details.

\subsection{Gaussian-Skeleton Distance and Joint Optimization}

L1 skeleton extraction, widely used in point cloud processing \cite{huang2013l1}, relies on point-to-skeleton distances but fails to account for key properties of 3D-GS, such as scale and uncertainty. To address this, Gaussian-to-Skeleton (G2D) distance is introduced to handle both sparse and dense Gaussians.  

Given a Gaussian \( h_i \) in the 3D-GS model \( \mathcal{H}^N \) and a set of skeleton points \( \mathcal{S} = \{\boldsymbol{q}_j\}_{j=1}^K \), the G2D distance is defined as:

\[
D\left( h_i,\mathcal{S} \right) =\min_{\boldsymbol{q}_j\in \mathcal{S}}\sum_{i\in \mathcal{G}_j}{\exp}\left( -\frac{1}{2}\left( \boldsymbol{q}_j-\boldsymbol{\mu }_i \right) ^{\top}\mathbf{\Sigma }_{i}^{-1}\left( \boldsymbol{q}_j-\boldsymbol{\mu }_i \right) \right) \tag{1}
\]
where \( \boldsymbol{q}_j \) is the \( j \)-th skeleton point, and \( \mathcal{G}_j \) is the set of Gaussians associated with this skeleton point. The Gaussian density is utilized to capture the spatial relationship between the skeleton point and the Gaussians. The first term aligns the skeleton with Gaussians, and the second enforces smoothness by penalizing curvature.

The skeleton initialization begins by clustering \( \mathcal{H} \) using DBSCAN, which identifies dense regions of Gaussians representing key structural features in the 3D scene \cite{ester1996density}. Each Gaussian is classified as a core point if its \(\epsilon\)-neighborhood, defined by Euclidean distance, contains at least \( \text{minPts} \) other Gaussians. Core points initiate clusters  \( \left\{ \mathcal{C}_i \right\} \), which are iteratively expanded by including neighboring Gaussians. The initial skeleton point set \(\mathcal{S}^0 = \{\boldsymbol{\mu}_{\mathcal{C}_1}, \dots, \boldsymbol{\mu}_{\mathcal{C}_K}\}\) is obtained after clustering, where \(K\) is the total number of clusters.

This weighted center weights denser Gaussians more in cluster formation, forming  the initial skeleton point set \(\mathcal{S}^0 = \{\boldsymbol{\mu}_{\mathcal{C}_1}, \dots, \boldsymbol{\mu}_{\mathcal{C}_K}\}\) is obtained through DBSCAN clustering, where \(K\) is the total number of clusters. DBSCAN’s adaptive density handling and noise tolerance ensure robust clustering, even with heterogeneous densities, overlaps, and outliers, providing a reliable initial skeleton without the need for a predefined number of clusters.

To refine the initial skeleton, a joint energy function \(E(\mathcal{S})\) combining data fitting and curvature regularization is used. The first term measures the fitting between the skeleton and the 3D Gaussians, while the second enforces smoothness by regularizing the skeleton’s curvature:

\[
E(\mathcal{S}) = \sum_{i=1}^N D(h_i, \mathcal{S}) + \lambda \sum_{\boldsymbol{q}_j \in \mathcal{S}} \|\nabla \boldsymbol{q}_j\|^2 \tag{2}
\]
where \(D(h_i, \mathcal{S})\) quantifies the distance between each Gaussian \(h_i\) and the skeleton points, and \(\lambda\) controls the trade-off between fitting and smoothness. The second term penalizes excessive curvature across the skeleton. The goal is to iteratively update the skeleton points \(\mathcal{S}\) to minimize this energy function, balancing accurate fitting to the Gaussian distributions with maintaining smoothness. In each iteration, Gaussian kernels are assigned to the nearest skeleton points using a greedy strategy that minimizes the distance. Once the assignments are fixed, the skeleton points are updated using gradient descent until the change in skeleton positions falls below a threshold $\varepsilon$, with adjacent skeleton points merged if their proximity exceeds a predefined threshold to ensure structural coherence. The curvature regularization term $\|\nabla \boldsymbol{q}_j\|^2$  approximates the local curvature at each point along the skeleton.

\subsection{Gaussian-Adaptive KPConv for Anisotropic Feature Extraction}

Conventional KPConv, which relies on Euclidean neighborhoods, is ineffective in capturing the anisotropic structure of Gaussians, limiting feature extraction in 3D-GS. To overcome this, a GA-KPConv framework is introduced, incorporating Mahalanobis distance-based neighborhood selection and ellipsoid-aware convolution kernels to enhance local geometric representation.

Traditional Euclidean-based neighborhoods are inadequate for processing anisotropic Gaussians, as they fail to adapt to varying spatial distributions. To address this, neighborhood \( N_{\boldsymbol{\mu }_i} \) is obtained using Mahalanobis distance, defined as:

\[
d_{\text{M}}\left( \boldsymbol{\mu }_i,\boldsymbol{\mu }_j;\mathbf{\Sigma }_i \right) =\sqrt{\left( \boldsymbol{\mu }_j-\boldsymbol{\mu }_i \right) ^{\top}\mathbf{\Sigma }_{i}^{-1}\left( \boldsymbol{\mu }_j-\boldsymbol{\mu }_i \right)} \tag{3}
\]

This adaptive measure expands the neighborhood in high-variance directions while contracting it in low-variance regions, selectively emphasizing intricate details while preserving smooth structures.

To enhance anisotropic feature extraction, a Gaussian-aware kernel weighting function is designed. Given a convolution center \( \boldsymbol{\mu }_i \) and its neighboring points \( \boldsymbol{\mu }_j \), the kernel weighting function is defined as:

\[
w_m\left( \boldsymbol{\mu }_j,\boldsymbol{\Sigma }_i,\boldsymbol{x}_k \right) =\exp \left( -\frac{1}{2}d_{k}^{\top}\boldsymbol{\Sigma }_{i}^{-1}d_k \right) \tag{4}
\]
where \(d_k = \bigl(\boldsymbol{\mu}_j - \boldsymbol{\mu}_i\bigr) - \boldsymbol{x}_k\) represents the relative displacement between the neighboring Gaussian center \(\boldsymbol{\mu}_j\) and the \(m\)-th kernel point \(\boldsymbol{x}_k \in \{\boldsymbol{x}_m\}_{m=1}^M\). In this work, the kernel points \(\{\boldsymbol{x}_m\}\) are fixed within a sphere of radius \(\sigma\), thus defining the kernel’s fundamental shape and size. Incorporating \(\boldsymbol{\Sigma}_i^{-1}\) dynamically morphs this spherical base into an ellipsoidal receptive field for each convolution center: larger covariance eigenvalues extend the kernel’s reach in the corresponding direction, while smaller eigenvalues shrink it, enabling orientation- and scale-aware feature extraction. The convolution operation is formulated as  

\[
F'\left( \boldsymbol{\mu }_i \right) =\sum_{\boldsymbol{\mu }_j\in N_{\boldsymbol{\mu }_i}}{w_m}\left( \boldsymbol{\mu }_j,\boldsymbol{\Sigma }_i,\boldsymbol{x}_k \right) f_j \tag{5}
\]
where \( f_j \) represents the input feature of the neighboring Gaussians \( \boldsymbol{\mu }_j \). The first two layers' output features \( F'\left( \boldsymbol{\mu }_i \right) _2 \) is used for local detail recognition, as they strongly respond to high-frequency structures within a small receptive field, with covariance-driven adaptation enhancing intricate details while suppressing smoother regions. A larger \( F'\left( \boldsymbol{\mu }_i \right) \) indicates richer geometric details such as high-curvature areas, sculpted features, and textures.

\subsection{Structure-Aware 3D-GS Fusion with Detail Preservation}

To achieve seamless 3D-GS scene fusion while preserving global structural consistency and local detail, a structure-aware strategy is introduced. Given two independently reconstructed 3D-GS models, \( \mathcal{H}_{\text{A}}^{N} \) and \( \mathcal{H}_{\text{B}}^{M} \), which have already undergone registration, fusion is performed to integrate them into a unified representation. The fusion process is carried out by aligning the models to a globally consistent skeleton \( \mathcal{S}_{\text{f}} \), constructed through the merging and refinement of individual skeletons \( \mathcal{S}_{\text{A}} \) and \( \mathcal{S}_{\text{B}} \) (Algorithm \ref{alg:3D-GSFusion}). Overlapping skeleton nodes are averaged to ensure structural continuity, while unmerged nodes are retained to preserve unique geometric features.

The scene is spatially partitioned into non-overlapping regions, \( \mathcal{R}_{\text{A}} \) and \( \mathcal{R}_{\text{B}} \), where Gaussians are directly retained, and an overlapping region \( \mathcal{R}_{\text{AB}} \), where fusion decisions are required. For each Gaussian in \( \mathcal{R}_{\text{AB}} \), a fusion score \( S_{\mathrm{tot}} \) is computed to determine its selection, balancing structural adherence, local detail fidelity, and spatial distribution consistency. The skeleton adherence score \( S_{\mathrm{ske}} \) quantifies how well a Gaussian aligns with \( \mathcal{S}_{\text{f}} \), ensuring structural coherence. The local detail score \( S_{\text{deta}} \) measures fine-scale geometric significance using feature responses, prioritizing high-frequency components. The scene center proximity score \( S_{\text{cen}} \) evaluates spatial bias relative to scene centers \( \boldsymbol{c}_{\text{A}} \) and \( \boldsymbol{c}_{\text{B}} \), normalized by characteristic radii \( R_{\text{A}} \) and \( R_{\text{B}} \), ensuring a balanced contribution from both reconstructions. The final selection is made by computing \( S_{\mathrm{tot}} \) as a weighted combination of these factors, with coefficients \( \alpha \), \( \beta \), and \( \gamma \), forming the optimized scene representation \( \mathcal{H}_{\text{f}} \).  

\begin{algorithm}[!t] \small
\caption{3D-GS Fusion Strategy}
\label{alg:3D-GSFusion}
\begin{algorithmic}
\ForAll{$\boldsymbol{q}_\text{a} \in \mathcal{S}_{\text{A}}$}
  \State $merged \gets \texttt{false}$
  \ForAll{$\boldsymbol{q}_\text{b} \in \mathcal{S}_{\text{B}}$}
    \If{$\|\boldsymbol{q}_\text{a} - \boldsymbol{q}_\text{b}\| < \varepsilon_{\text{skel}}$}
      \State $\mathcal{S}_{\text{f}} \gets \mathcal{S}_{\text{f}} \cup \{(\boldsymbol{q}_\text{a}+\boldsymbol{q}_\text{b})/2\}$ \Comment{Merge close nodes}
      \State Mark $\boldsymbol{q}_\text{b}$ as merged, $merged \gets \texttt{true}$; \textbf{break}
    \EndIf
  \EndFor
  \If{not $merged$}
    \State $\mathcal{S}_{\text{f}} \gets \mathcal{S}_{\text{f}} \cup \{\boldsymbol{q}_\text{a}\}$ \Comment{Retain unmerged node from $\mathcal{S}_{\text{A}}$}
  \EndIf
\EndFor
\ForAll{$\boldsymbol{q}_\text{b} \in \mathcal{S}_{\text{B}}$}
  \If{ $\boldsymbol{q}_\text{b}$ is not marked as merged}
    \State $\mathcal{S}_{\text{f}} \gets \mathcal{S}_{\text{f}} \cup \{\boldsymbol{q}_\text{b}\}$ \Comment{Add remaining nodes from $\mathcal{S}_{\text{B}}$}
  \EndIf
\EndFor

\State $\mathcal{R}_{\text{AB}}\gets \bigl\{ \left( \boldsymbol{\mu }_{i}^{\text{A}},\boldsymbol{\mu }_{j}^{\text{B}} \right) \mid |\boldsymbol{\mu }_{i}^{\text{A}}-\boldsymbol{\mu }_{j}^{\text{B}}|\le \varepsilon _{\text{overlap}} \bigr\} $
\ForAll{ $(\boldsymbol{\mu }_{i}^{\text{A}},\boldsymbol{\mu }_{j}^{\text{B}})\in \mathcal{R}_\text{AB}$}
  \ForAll{ $\boldsymbol{\mu } \in \{\boldsymbol{\mu }_i^\text{A},\boldsymbol{\mu }_j^\text{B}\}$}
    \State $S_{\mathrm{ske}}(\boldsymbol{\mu })\gets \dfrac{1}{1+\delta \min_{\boldsymbol{q}\in\mathcal{S}_{\text{f}}}\|\boldsymbol{\mu }-\boldsymbol{q}\|}$
    \State $S_{\text{deta}}\left( \boldsymbol{\mu } \right) \gets \dfrac{F'(\boldsymbol{\mu })-F_{\min}}{F_{\max}-F_{\min}}$
    \State $\bigl(\boldsymbol{c}(\boldsymbol{\mu }),R(\boldsymbol{\mu })\bigr)\!\gets (\boldsymbol{c}_{\text{A}},R_{\text{A}})\;\textbf{if}\;\boldsymbol{\mu }\in\mathcal{H}_{\text{A}}^{N};\textbf{else}\;(\boldsymbol{c}_{\text{B}},R_{\text{B}})$
    \State $S_{\text{cen}}\left( \boldsymbol{\mu } \right)\gets 1 - \dfrac{\|\boldsymbol{\mu } - \boldsymbol{c}(\boldsymbol{\mu })\|}{R(\boldsymbol{\mu })}$
    \State $S_{\mathrm{tot}}(\boldsymbol{\mu })\!\gets \alpha\,S_{\mathrm{ske}}(\boldsymbol{\mu })\,+\,\beta\,S_{\text{deta}}\left( \boldsymbol{\mu } \right) \,+\,\gamma\,S_{\mathrm{ske}}(\boldsymbol{\mu })$
  \EndFor
    \State $\mathcal{H}_{\text{f}} \gets \boldsymbol{\mu }_i^\text{A}$ \textbf{if} $S_{\mathrm{tot}}(\boldsymbol{\mu }_i^\text{A}) \ge S_{\mathrm{tot}}(\boldsymbol{\mu }_j^\text{B}) \;\textbf{or}\; \max\bigl(S_{\mathrm{ske}}(\boldsymbol{\mu }_i^\text{A}),\,S_{\text{deta}}(\boldsymbol{\mu }_i^\text{A})\bigr) \ge \tau$ \textbf{else} $\boldsymbol{\mu }_j^\text{B}$
    \EndFor
    \end{algorithmic}
\end{algorithm}

\section{EXPERIMENTS}
\subsection{Experiment Setup}

To evaluate the proposed method, we conduct experiments on the ScanNet-GSReg \cite{chang2024gaussreg} and our Coord dataset, both of which contain varying point densities and structural complexities. All methods were implemented in PyTorch and Open3D and tested on an Intel Xeon processor with an NVIDIA RTX 3090 GPU. 

Performance was measured using Hausdorff distance and curvature deviation (Curv. Dev.) for skeleton accuracy, feature contrast and matching accuracy for local detail evaluation, and rotation registration error (RRE), translation registration error (RTE), scale registration error (RSE), and computation time for registration performance. Skeleton extraction was evaluated by comparing curvature deviation, connectivity (Conn.), and computation time (Time) between the proposed GA-L1 and L1-based methods. Curvature deviation quantifies the smoothness of the extracted skeleton by measuring the mean absolute curvature difference along the structure, while connectivity (Conn.) assesses the graph connectivity by computing the ratio of valid skeleton edges to the total expected connections within clustered regions. The registration pipeline integrated the proposed feature extraction method with GeoTransformer and was benchmarked against GaussReg-NF (GaussReg w/o. fine), which employs standard KPConv for feature extraction. Finally, the fusion strategy was compared between GaussReg-NF and the proposed method, using PSNR, SSIM, and LPIPS to evaluate reconstruction quality \cite{kerbl20233d}.

\subsection{Registration Enhancement via GA-KPConv Feature Extraction}

\begin{figure*}[]
\centering
\includegraphics[width=0.7\linewidth]{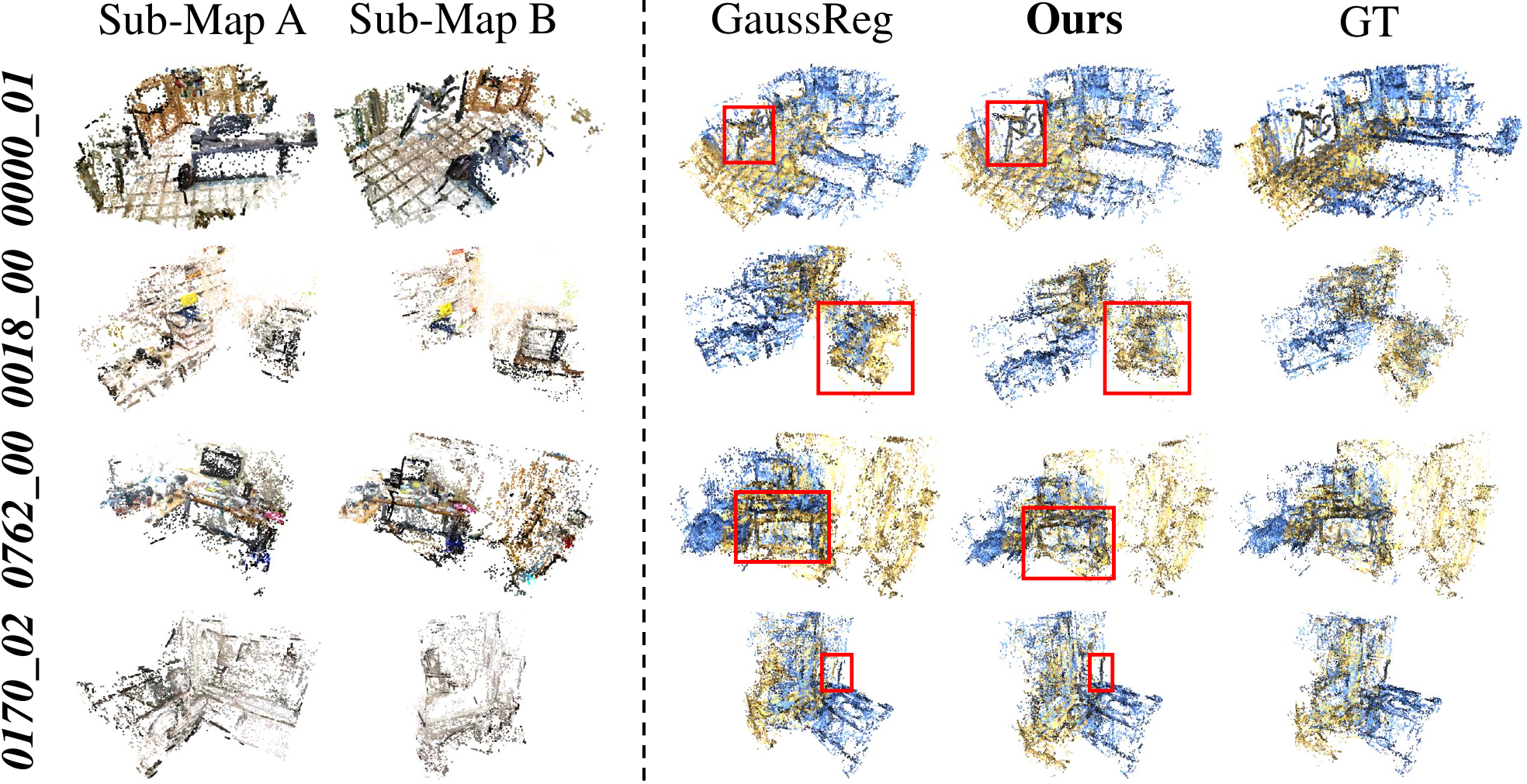} 
\captionsetup{font=small}
\caption{Qualitative comparison of registration results on the ScanNet-GSReg dataset.}
\label{fig:reg_comparison}
\end{figure*}

Table~\ref{tab:reg_comparison} shows that the proposed method consistently outperforms GaussReg-NF in registration accuracy. In particular, Scene 0089-01 exhibits the most pronounced improvement, with a 41.9\% reduction in RRE and an 82.3\% decrease in RTE, highlighting the robustness of the proposed method in challenging environments with varying densities and structural anisotropies. Although the method incurs a 21.4\% increase in processing time, the enhancement in registration accuracy stems from the covariance-aware feature extraction in GA-KPConv, which enables more robust alignment in complex geometric regions. While GaussReg-NF achieves faster inference in some cases, the proposed approach consistently delivers superior structural alignment and registration accuracy, particularly in scenes with intricate geometric variations and anisotropic distributions. These results validate its effectiveness in enhancing multi-view 3D-GS registration while maintaining spatial coherence.

\begin{table*}[htbp]
\centering
\setlength{\tabcolsep}{4pt}
\caption{Quantitative comparison of registration results on the ScanNet-GSReg dataset. The ellipsis indicates additional tested but omitted results, and Avg. represents the mean computed over the ScanNet-GSReg test set.}\begin{tabular}{@{}c | c | c c c c c c c c c c c c | c@{}}
\toprule
\textbf{Methods} & \textbf{Metrics} & \textbf{0000-01} & \textbf{0018-00} & \textbf{0089-01} & \textbf{0170-02} & \textbf{0175-00} & \textbf{0727-00} & \textbf{0740-00} & \textbf{0756-00} & \textbf{0771-00} & \textbf{0796-00} & \textbf{0806-00} & \dots & \textbf{Avg.} \\
\midrule
\multirow{4}{*}{GaussReg-NF}  
  & RRE↓(\%)  & 2.150  & \textbf{1.122}  & 6.153  & 3.452  & 4.352  & 3.624  & 2.796  & 3.277  & 2.414  & 3.488  & 4.266  & \dots & 3.418 \\
  & RTE↓(m)   & \textbf{0.072}  & 0.054  & 0.062  & 0.053  & 0.059  & 0.069  & 0.054  & 0.054  & 0.057  & \textbf{0.052}  & 0.038  & \dots & 0.060 \\
  & RSE↓(\%)  & 0.021  & 0.051  & 0.030  & \textbf{0.045}  & \textbf{0.016}  & 0.018  & 0.025  & 0.017  & 0.030  & 0.049  & 0.023  & \dots & 0.029 \\
  & Time↓(s)  & \textbf{5.105}  & 6.101  & 5.314  & \textbf{2.522}  & \textbf{3.674}  & \textbf{4.050}  & \textbf{5.856}  & \textbf{6.347}  & 6.228  & \textbf{5.825}  & \textbf{4.314}  & \dots & \textbf{4.151} \\
\midrule
\multirow{4}{*}{Ours}  
  & RRE↓(\%)  & \textbf{1.923}  & 1.254  & \textbf{3.568}  & \textbf{1.753}  & \textbf{2.624}  & \textbf{1.562}  & \textbf{2.374}  & \textbf{3.043}  & \textbf{1.793}  & \textbf{2.435}  & \textbf{3.294}  & \dots & \textbf{2.595} \\
  & RTE↓(m)   & 0.075  & \textbf{0.046}  & \textbf{0.011}  & \textbf{0.036}  & \textbf{0.054}  & \textbf{0.063}  & \textbf{0.036}  & \textbf{0.032}  & \textbf{0.048}  & 0.051  & \textbf{0.034}  & \dots & \textbf{0.045} \\
  & RSE↓(\%)  & \textbf{0.018}  & \textbf{0.040}  & \textbf{0.018}  & 0.056  & \textbf{0.016}  & \textbf{0.017}  & \textbf{0.022}  & \textbf{0.013}  & \textbf{0.015}  & \textbf{0.017}  & \textbf{0.012}  & \dots & \textbf{0.013} \\
  & Time↓(s)  & 5.751  & \textbf{5.621}  & \textbf{4.892}  & 4.320  & 6.099  & 4.413  & 4.868  & 6.397  & \textbf{3.603}  & 6.318  & 4.666  & \dots & 5.041 \\
\bottomrule
\end{tabular}
\label{tab:reg_comparison}
\end{table*}

To assess the impact of Gaussian-adaptive feature extraction on registration, Fig.~\ref{fig:reg_comparison} presents the registration results of the proposed method and GaussReg-NF on the ScanNet-GSReg dataset. Compared to GaussReg-NF, the proposed method achieves more precise alignment, particularly in high-curvature and anisotropic regions, where standard feature extractors struggle. The improvement stems from GA-KPConv’s Mahalanobis distance-based neighborhood selection and ellipsoid-aware convolution, which enhance feature distinctiveness in geometrically and radiometrically complex areas. As a result, the proposed method reduces registration errors and misalignment artifacts, ensuring more accurate transformations in challenging reconstruction scenarios.

\begin{figure}[htbp]
\centering
\includegraphics[width=\linewidth]{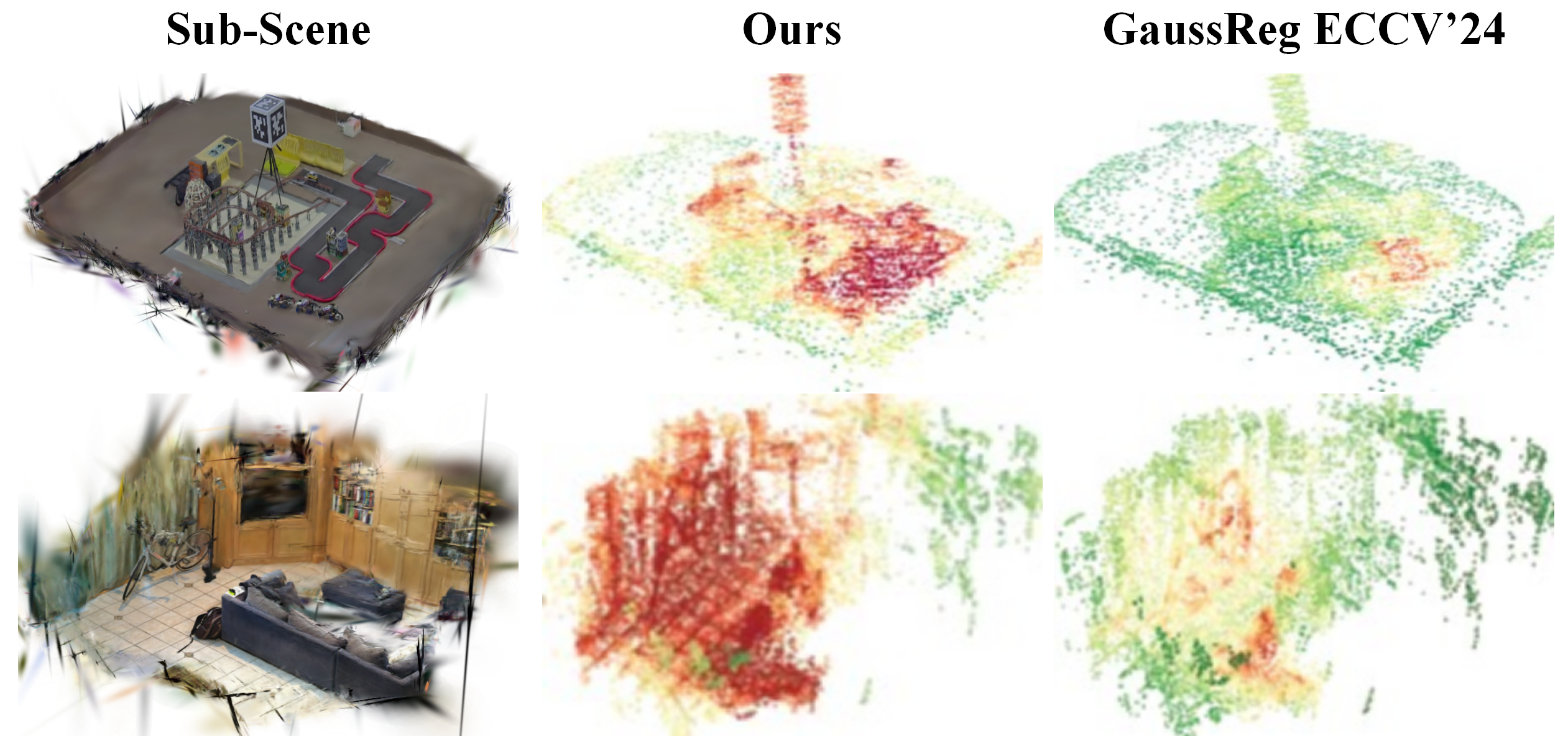} 
\captionsetup{font=small}
\caption{\textbf{Visualization of feature extraction results on 3D-GS scenes.} Warmer colors (red) indicate richer local details.}
\label{fig:feature_response}
\end{figure}

\begin{figure*}[b]
    \centering
    \begin{subfigure}{0.15\linewidth}
        \includegraphics[width=\linewidth]{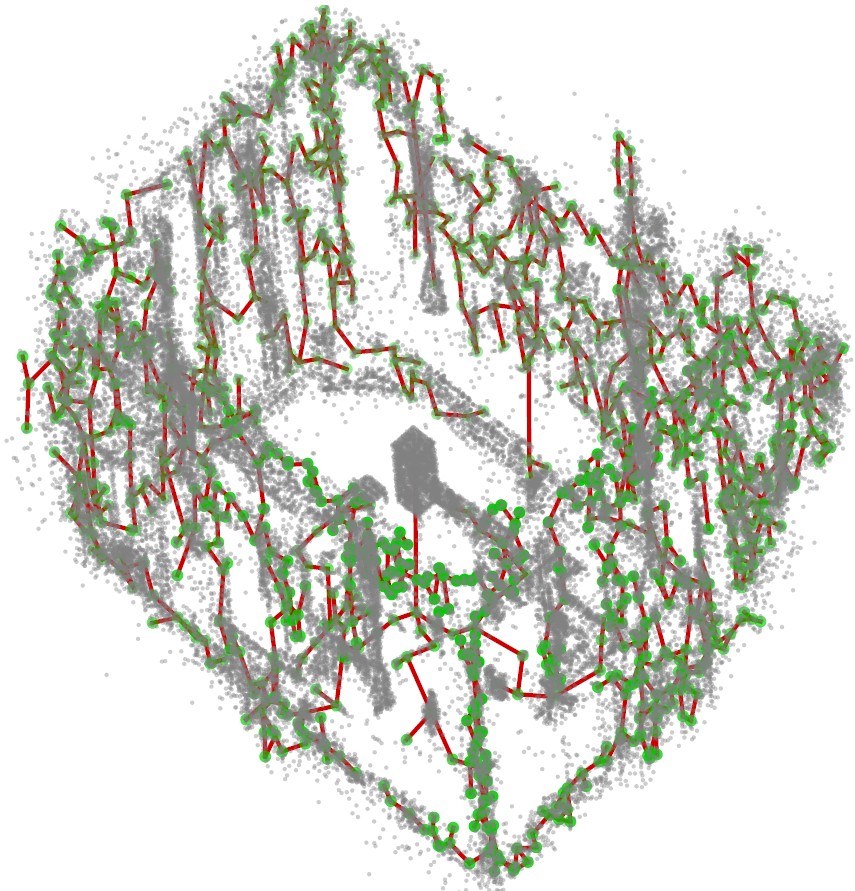}
        \captionsetup{font=small}
        \caption*{Iteration 0}
    \end{subfigure}
    \centering
    \begin{subfigure}{0.15\linewidth}
        \includegraphics[width=\linewidth]{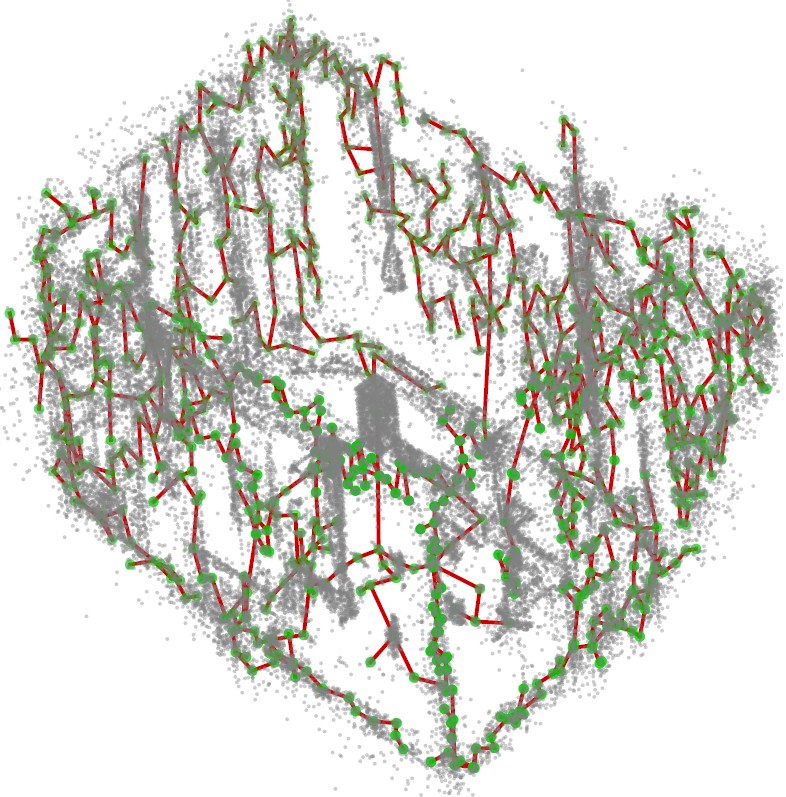}
        \captionsetup{font=small}
        \caption*{Iteration 10}
    \end{subfigure}
    \centering
    \begin{subfigure}{0.15\linewidth}
        \includegraphics[width=\linewidth]{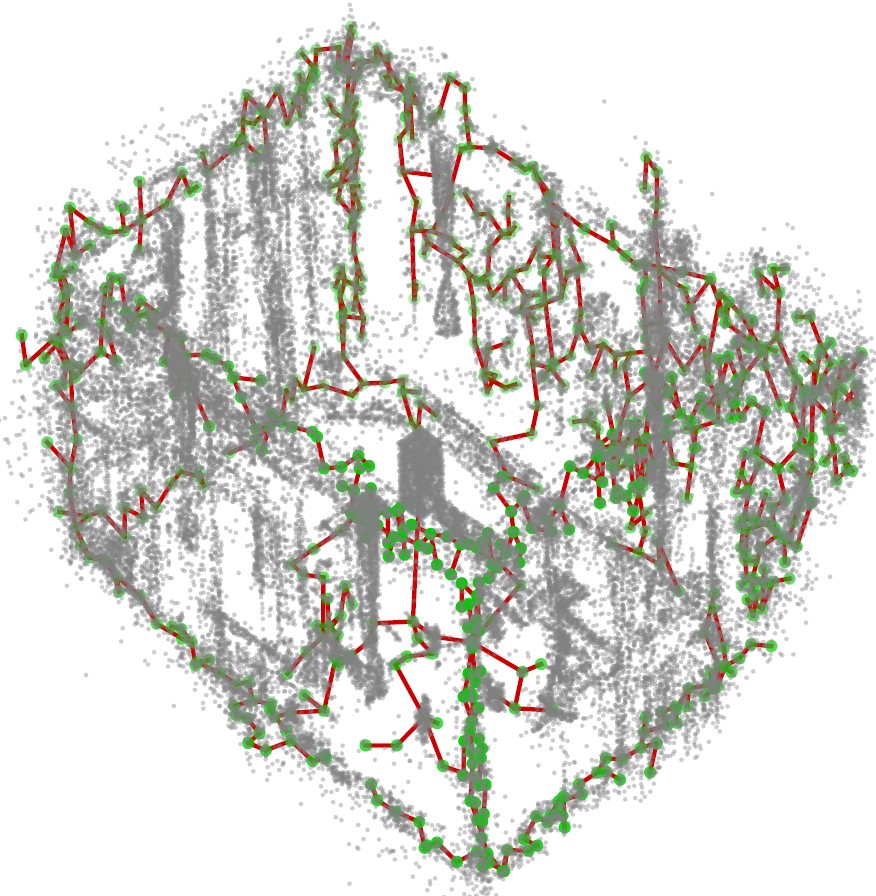}
        \captionsetup{font=small}
        \caption*{Iteration 30}
    \end{subfigure}
    \centering
    \begin{subfigure}{0.15\linewidth}
        \includegraphics[width=\linewidth]{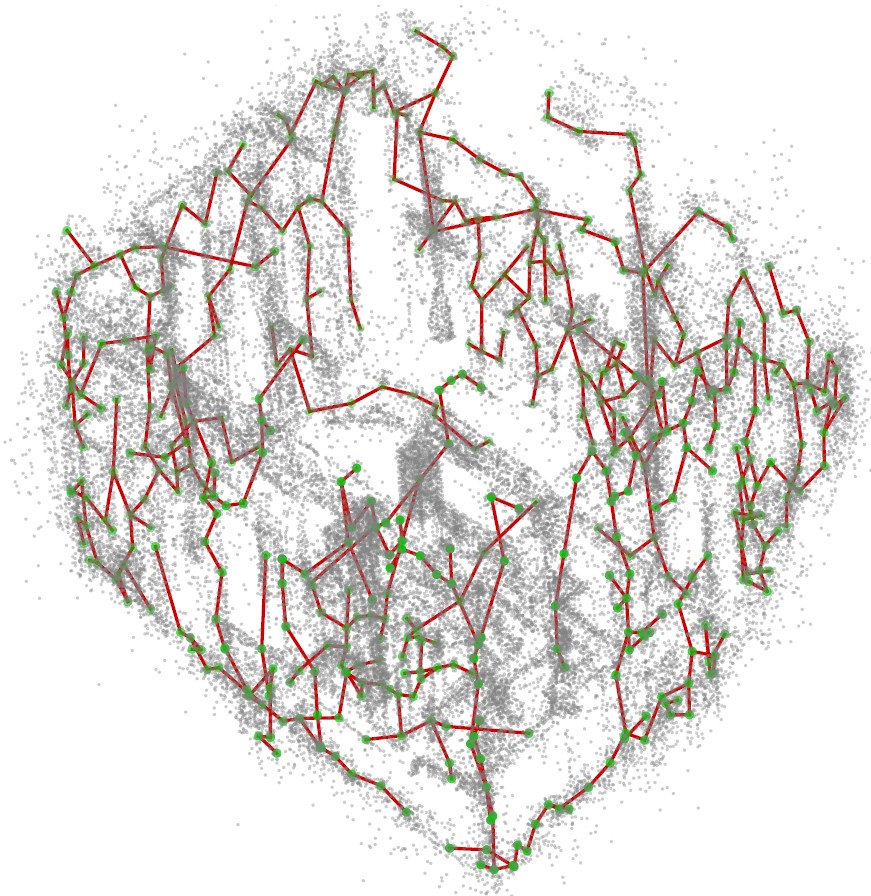}
        \captionsetup{font=small}
        \caption*{Iteration 90}
    \end{subfigure}
    \centering
    \begin{subfigure}{0.15\linewidth}
        \includegraphics[width=\linewidth]{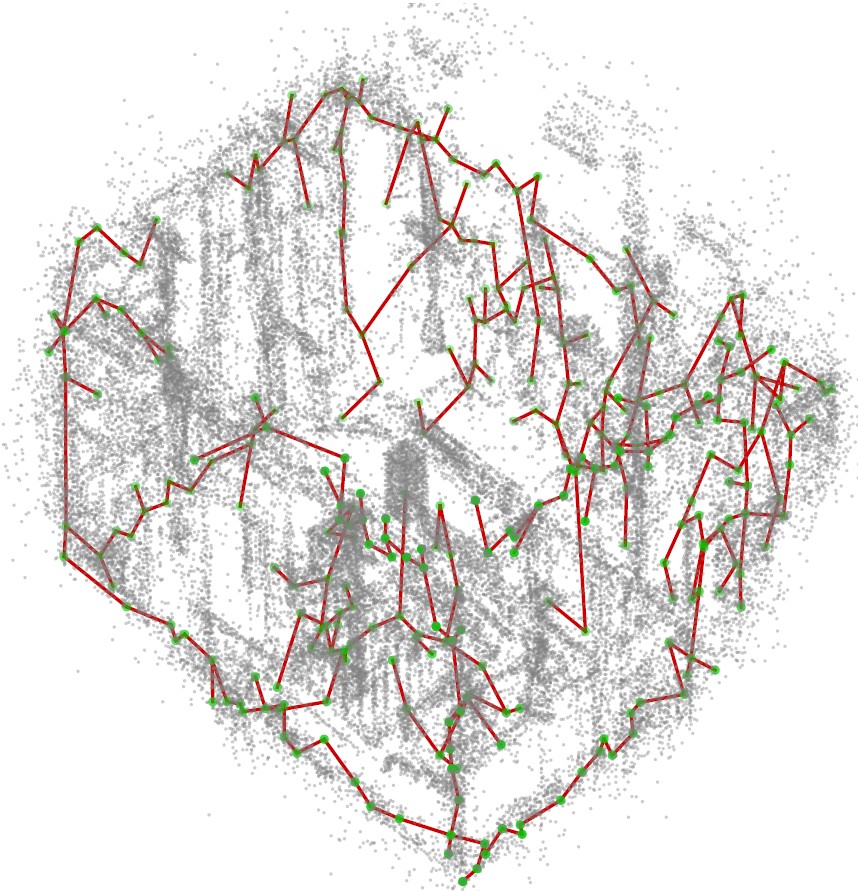}
        \captionsetup{font=small}
        \caption*{Iteration 150}
    \end{subfigure}
    \centering
    \begin{subfigure}{0.15\linewidth}
        \includegraphics[width=\linewidth]{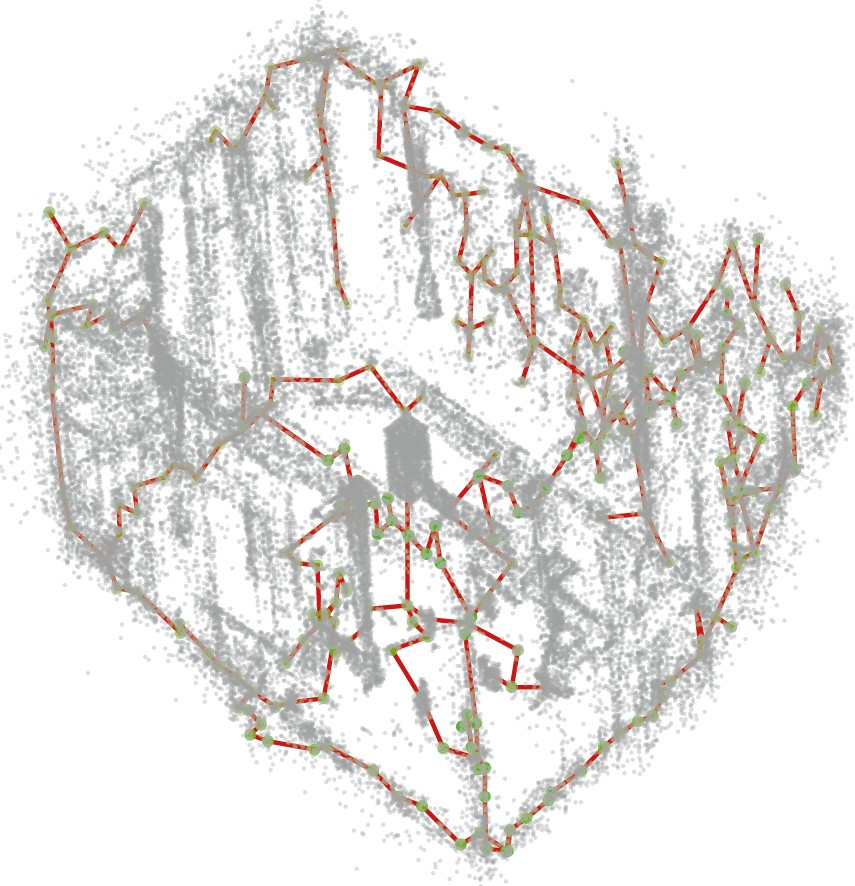}
        \captionsetup{font=small}
        \caption*{Iteration 400}
    \end{subfigure}

    \centering
    \begin{subfigure}{0.15\linewidth}
        \includegraphics[width=\linewidth]{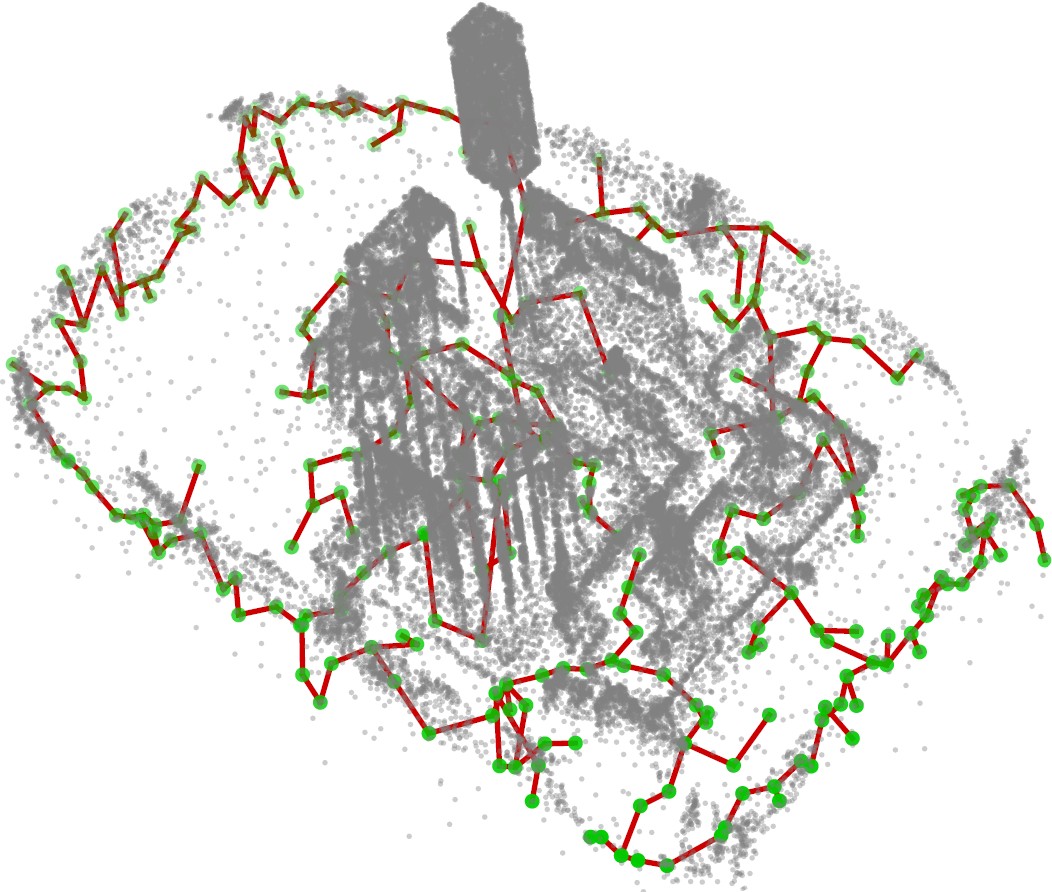}
        \captionsetup{font=small}
        \caption*{Iteration 0}
    \end{subfigure}
    \centering
    \begin{subfigure}{0.15\linewidth}
        \includegraphics[width=\linewidth]{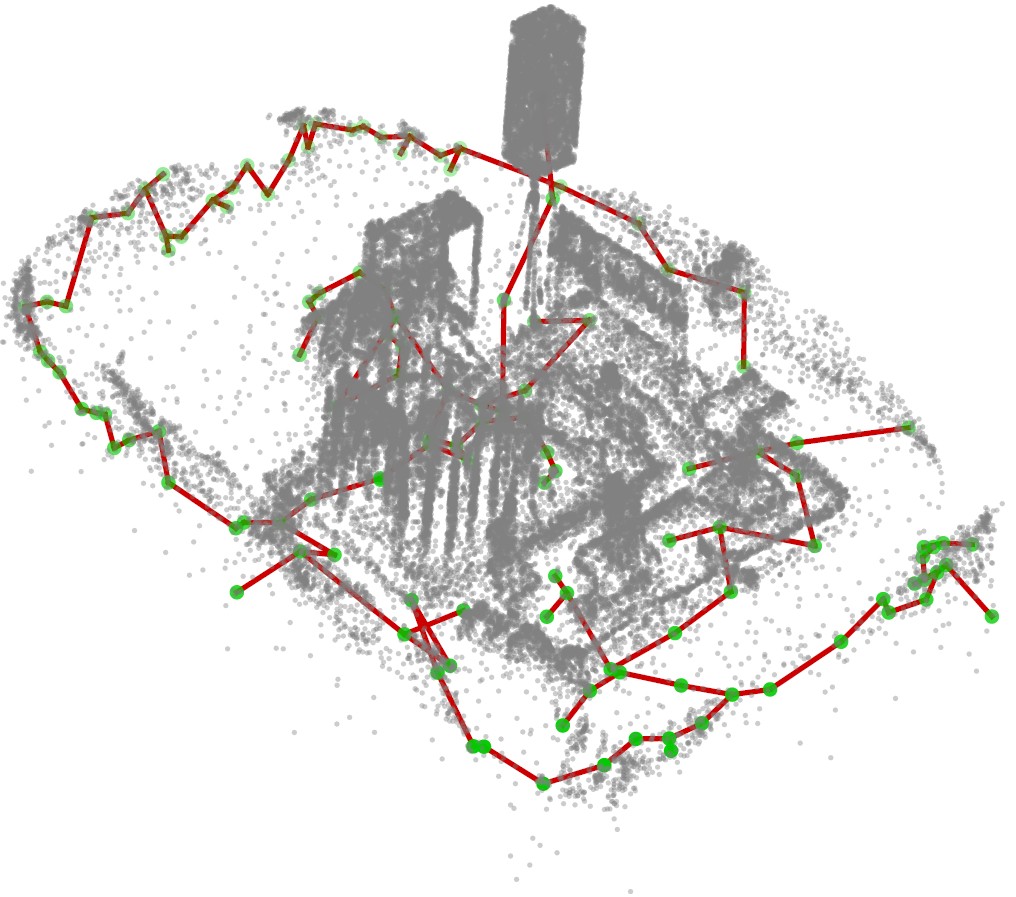}
        \captionsetup{font=small}
        \caption*{Iteration 2}
    \end{subfigure}
    \centering
    \begin{subfigure}{0.15\linewidth}
        \includegraphics[width=\linewidth]{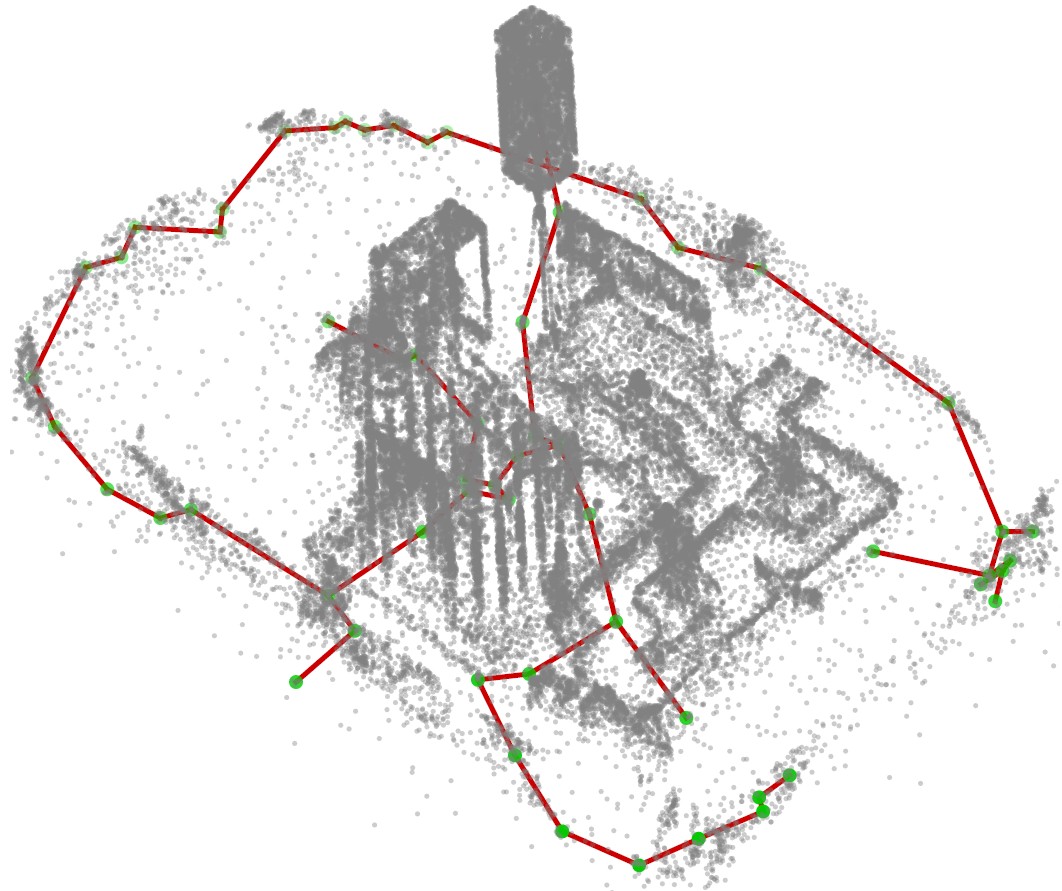}
        \captionsetup{font=small}
        \caption*{Iteration 5}
    \end{subfigure}
    \centering
    \begin{subfigure}{0.15\linewidth}
        \includegraphics[width=\linewidth]{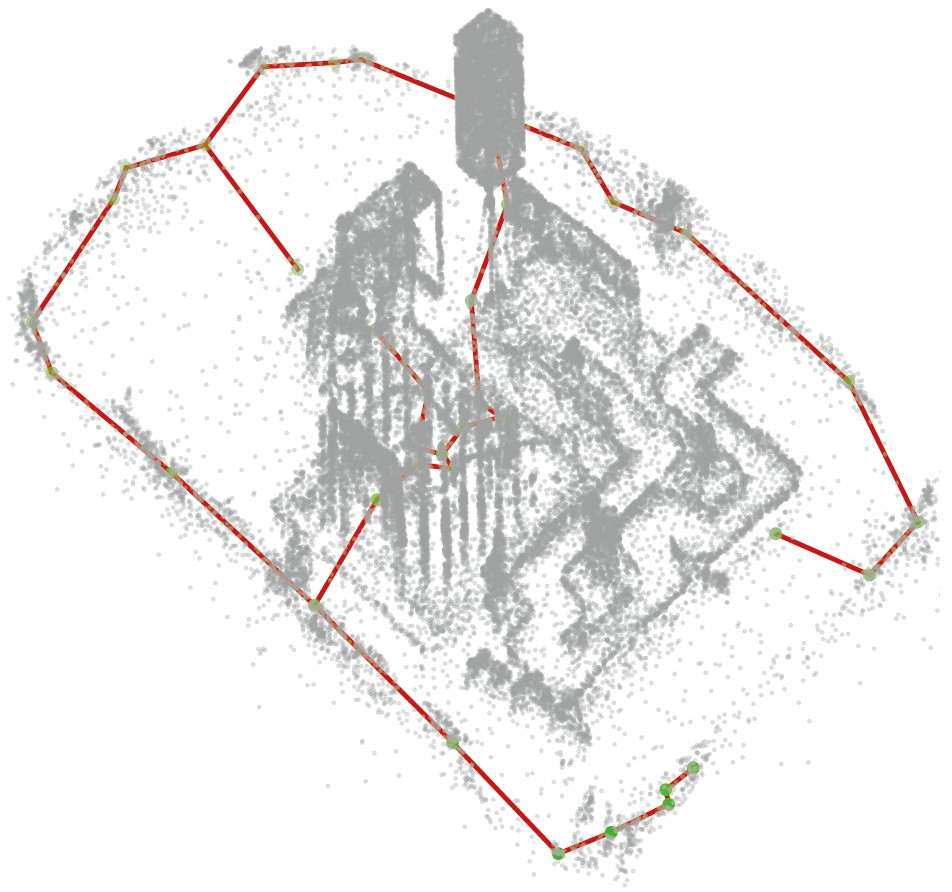}
        \captionsetup{font=small}
        \caption*{Iteration 10}
    \end{subfigure}
    \centering
    \begin{subfigure}{0.15\linewidth}
        \includegraphics[width=\linewidth]{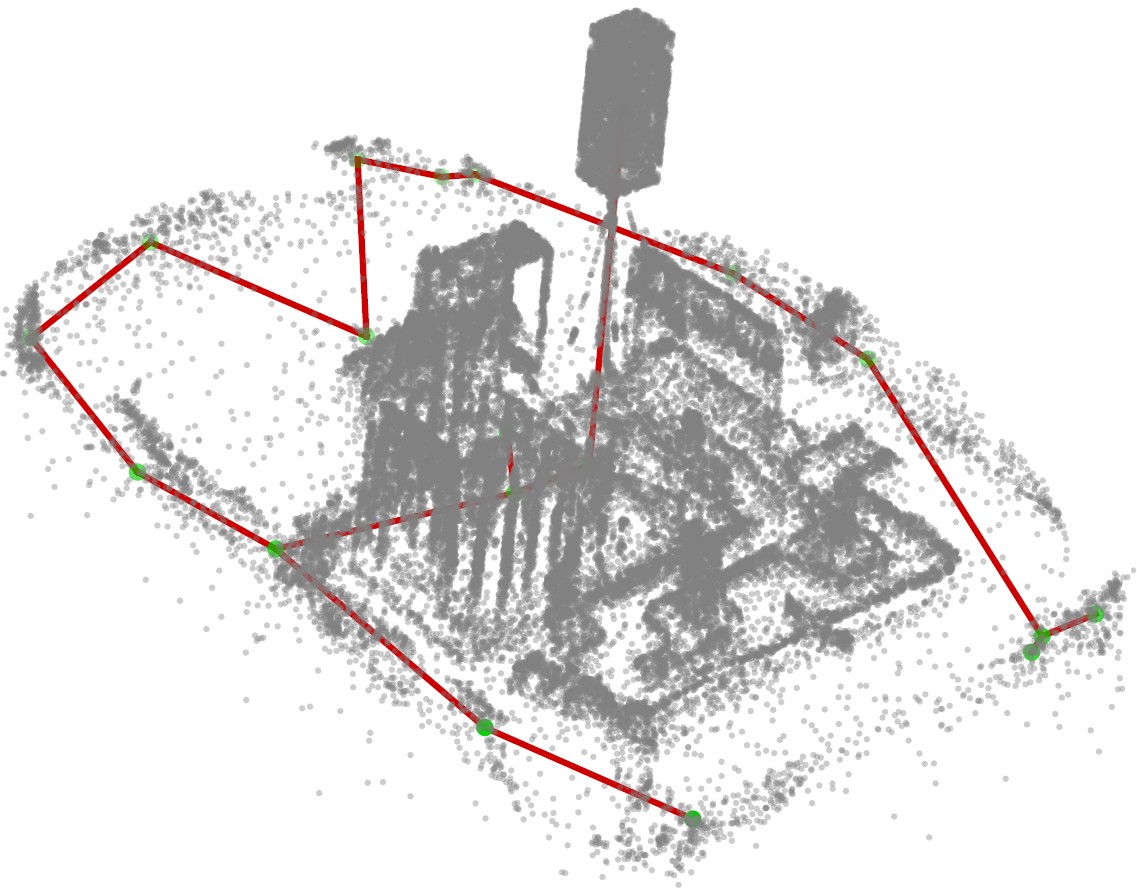}
        \captionsetup{font=small}
        \caption*{Iteration 110}
    \end{subfigure}
    \centering
    \begin{subfigure}{0.15\linewidth}
        \includegraphics[width=\linewidth]{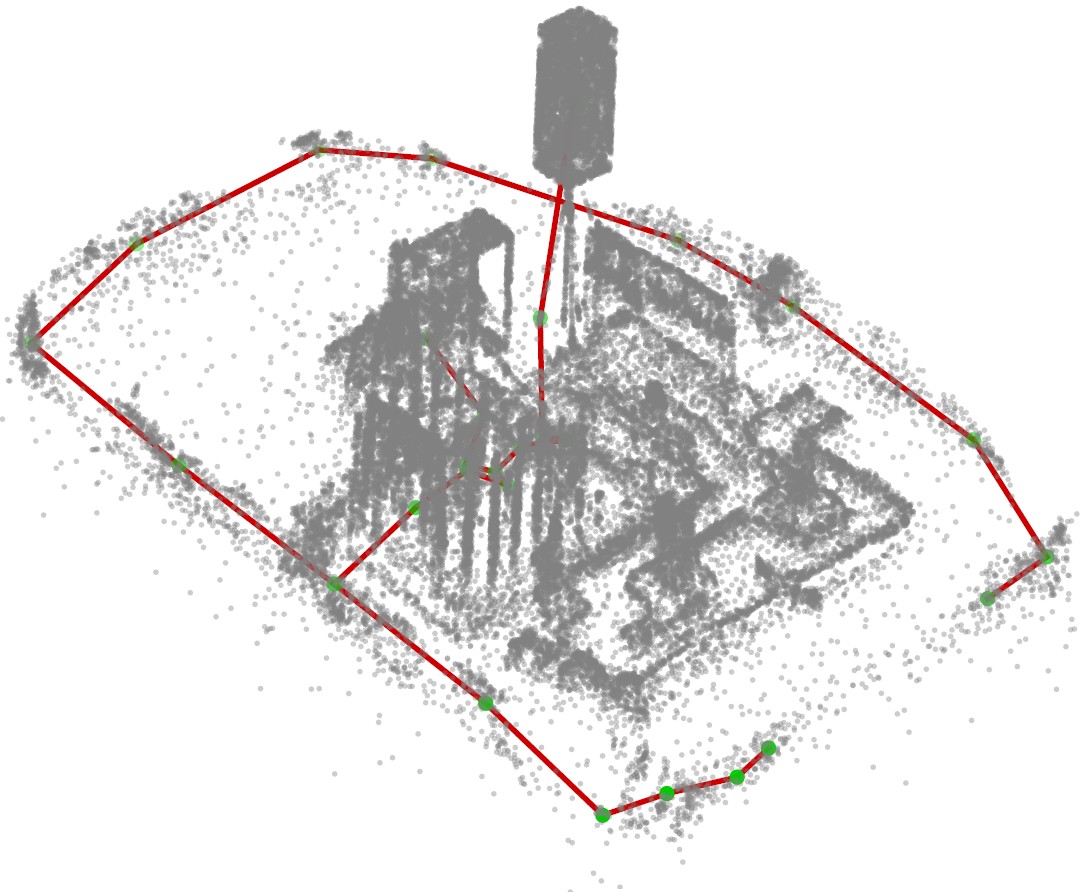}
        \captionsetup{font=small}
        \caption*{Iteration 300}
    \end{subfigure}
    \captionsetup{font=small}
    \caption{\textbf{Iterative optimization of the GA-L1 skeleton.} The first row depicts the stepwise refinement of the drone view skeleton, while the second row shows the optimization process of the vehicle view skeleton.}
    \label{fig:optimization_process}
\end{figure*}

To further explain the registration accuracy improvements, we analyze the feature extraction effectiveness of GA-KPConv and GaussReg-NF across 3D-GS scenes with varying Gaussian densities and anisotropic properties. Fig.~\ref{fig:feature_response} compares the feature maps extracted by the two approaches across a variety of scenes. 

\subsection{Skeleton Extraction and Local Detail Recognition for 3D-GS Fusion}

Beyond feature extraction, we evaluate the proposed G2D distance optimization in terms of skeleton accuracy, smoothness, and robustness across diverse 3D-GS datasets. The evaluation unfolds in three stages: (1) initialization using DBSCAN clustering, (2) optimization-based refinement, and (3) comparison with L1 skeleton extraction.

The optimization process refines the skeleton from both vehicle and drone views, as illustrated in Fig.~\ref{fig:optimization_process}. In the drone view , skeleton points iteratively adjust to better capture elevated structures and large-scale anisotropies. Early iterations exhibit significant shifts along major structural edges, while later iterations refine connectivity around high-curvature areas, such as rooftop boundaries, and progressively merge nearby skeleton points to eliminate redundancy and enhance structural coherence. By the final stage, the skeleton becomes more cohesive, accurately tracing prominent scene contours. In contrast, the vehicle view optimization converges more rapidly due to fewer elevation disparities. The skeleton stabilizes around structural supports, including building columns and wall edges, with subsequent iterations refining corner junctions and ensuring continuity along ground-level features. 

A direct comparison of G2D-based skeleton optimization  and L1 skeleton extraction under varying DBSCAN parameters is presented in Table~\ref{tab:skeleton_comparison}. Across all tested settings, GA-L1 consistently achieves lower curvature deviation, indicating smoother and more coherent skeletons. For example, at \(\epsilon = 1.0\) and \(\text{min\_Pts} = 8\), GA-L1 reduces curvature deviation to 3.5\%, compared to 7.9\% with L1, nearly a twofold improvement in smoothness. The connectivity metric also shows GA-L1 producing more structurally continuous skeletons, rising from 0.92 with L1 to 0.97 in the same parameter setting. Although GA-L1 incurs a moderate increase in computation time (e.g., 16.1\,s vs. 14.6\,s), this overhead is offset by its enhanced structural fidelity. Further evaluations measuring the Hausdorff distance and mean per-point registration error (not reported in the table) reveal similar trends, with GA-L1 consistently outperforming L1-based approaches, particularly in scenes with complex or anisotropic Gaussian distributions. These findings highlight the effectiveness of incorporating Gaussian covariance information into the optimization process, leading to a more robust and precise skeleton representation.

\begin{table}[htbp]
\centering
\setlength{\tabcolsep}{4pt}
\caption{Comparison of skeleton optimization results under different DBSCAN parameters.}
\begin{tabular}{@{}c c c c c@{}}
\toprule
\textbf{DBSCAN Params} & \textbf{Method} & \textbf{Curv. Dev↓(\%)} & \textbf{Conn.↑} & \textbf{Time↓(s)} \\
\midrule
\multirow{2}{*}{\(\epsilon = 0.5\), minPts = 4}  
  & GA-L1 & \textbf{5.62} &\textbf{ 0.924} & 12.51 \\
  & L1    & 9.71 & 0.887 & \textbf{10.18} \\
\multirow{2}{*}{\(\epsilon = 0.7\), minPts = 6}  
  & GA-L1 & \textbf{4.07} & \textbf{0.956} & 14.32 \\
  & L1    & 8.22 & 0.901 & \textbf{12.69} \\
\multirow{2}{*}{\(\epsilon = 1.0\), minPts = 8}  
  & GA-L1 & \textbf{3.54} & \textbf{0.974} & 16.12 \\
  & L1    & 7.90 & 0.920 & \textbf{14.56} \\
\bottomrule
\end{tabular}
\label{tab:skeleton_comparison}
\end{table}

To assess the impact of skeleton-guided fusion we register and fuse two 3D-GS sub-maps and compare the results against GaussReg, a baseline method that fuses Gaussians based on scene-center proximity. After GA-KPConv extracts local features, a GeoTransformer module estimates the relative rotation and translation transformation between partial reconstructions, ensuring robust alignment before fusion.

\begin{figure}[]
\centering
\includegraphics[width=\linewidth]{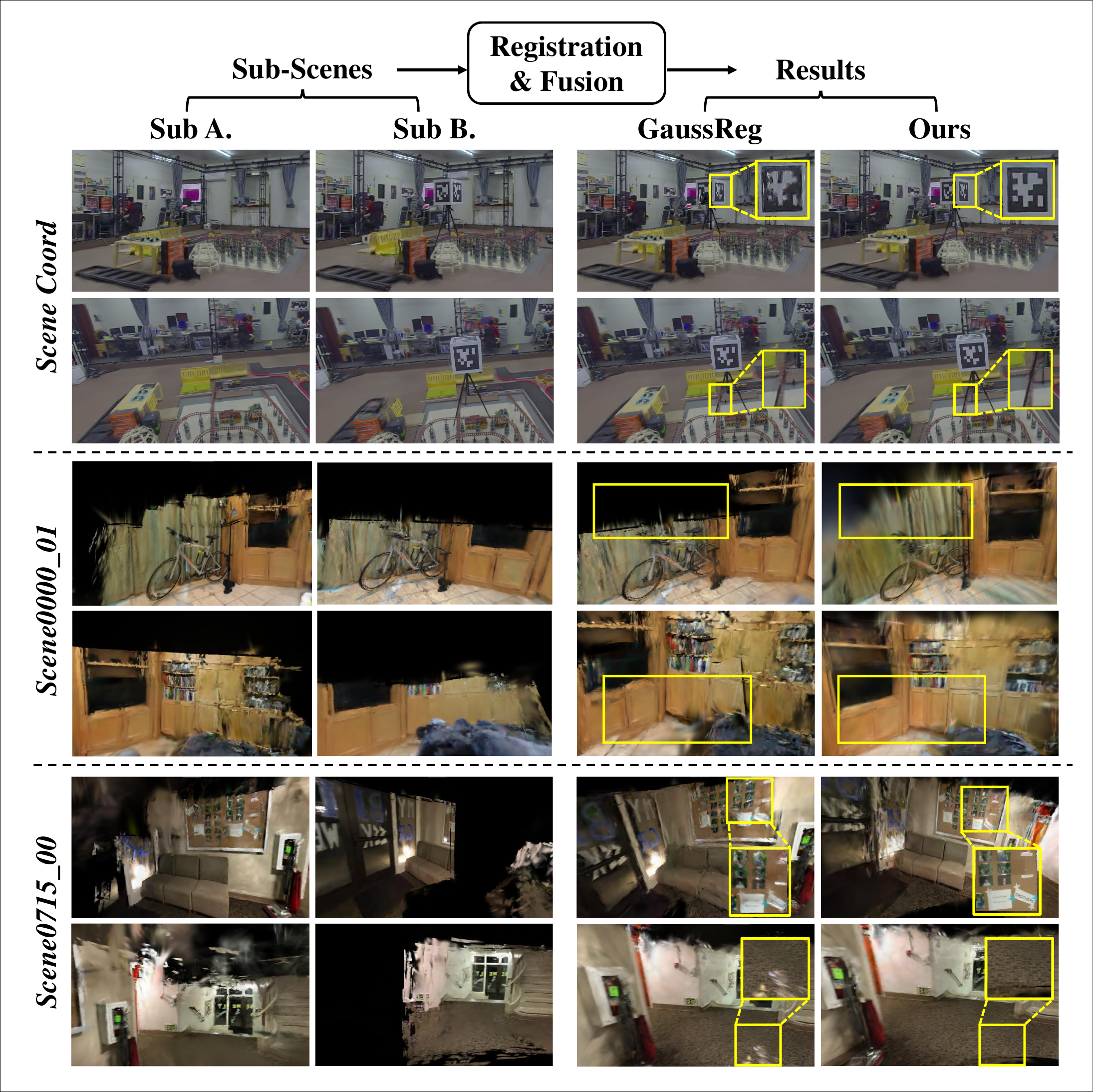} 
\captionsetup{font=small}
\caption{Qualitative comparison of fusion results on the ScanNet-GSReg dataset. Each scene is shown from two different viewpoints to provide a comprehensive comparison.}
\label{fig:fusion_comparison}
\end{figure}

\begin{table*}[]
\centering
\setlength{\tabcolsep}{4pt}
\caption{Quantitative comparison of registration results on the ScanNet-GSReg dataset. Sub-map B is registered to the coordinate system of sub-map A, and rendering metrics are computed using A’s camera poses for image generation.}
\begin{tabular}{@{}c | c | c c c c c c c c c c c c | c@{}}
\toprule
\textbf{Methods} & \textbf{Metrics} & \textbf{0000-01} & \textbf{0018-00} & \textbf{0715-00} & \textbf{0727-00} & \textbf{0740-00} & \textbf{0756-00} & \textbf{0763-00} & \textbf{0779-00} & \textbf{0790-00} & \textbf{0800-00} & \textbf{0806-00} & \dots  & \textbf{Avg.}\\
\midrule
\multirow{3}{*}{GaussReg}  
  & PSNR↑  & \textbf{22.781} & 22.402 & 17.762 & 21.498 & \textbf{22.832} & \textbf{21.728} & 16.432 & 21.324 & 20.478 & \textbf{20.432} & 20.308 & \dots & 20.270 \\
  & SSIM↑  & 0.840 & 0.862 & 0.717 & 0.779 & 0.783 & 0.778 & 0.738 & \textbf{0.790} & 0.804 & \textbf{0.801} & \textbf{0.826} & \dots & 0.778 \\
  & LPIPS↓ & 0.321 & 0.334 & 0.399 & 0.387 & 0.332 & 0.346 & 0.401 & \textbf{0.364} & \textbf{0.343} & 0.341 & 0.387& \dots & 0.360 \\
\midrule
\multirow{3}{*}{Ours}  
  & PSNR↑  & 22.562 & \textbf{22.860} & \textbf{27.871} & \textbf{23.803} & 20.082 & 21.165 & \textbf{24.491} & \textbf{22.493} & \textbf{22.853} & 20.126 & \textbf{20.915} & \dots & \textbf{25.212} \\
  & SSIM↑  & \textbf{0.859} & 0.856 & \textbf{0.877} & \textbf{0.806} & \textbf{0.82} & \textbf{0.788} & \textbf{0.864} & 0.789 & \textbf{0.816} & 0.789 & 0.780 & \dots & \textbf{0.868} \\
  & LPIPS↓ & \textbf{0.301} & \textbf{0.291} & \textbf{0.309} & \textbf{0.306} & \textbf{0.318} & 0.379 & \textbf{0.347} & 0.389 & 0.352 & \textbf{0.325} & \textbf{0.363} & \dots & \textbf{0.305} \\
\bottomrule
\end{tabular}
\label{tab:fusion_comparison}
\end{table*}

Table~\ref{tab:fusion_comparison} demonstrates the effectiveness of the proposed skeleton-guided 3D-GS fusion, particularly in complex and sparsely observed environments. In Scene 0715-00, where occlusions cause missing details, the proposed method achieves a 10.11 dB PSNR gain and a 22.3\% LPIPS reduction, preserving structural integrity by ensuring Gaussians are retained in regions with high local detail significance, preventing gaps and incomplete surfaces.

Scene 0763-00, characterized by intricate geometric structures, exhibits an 8.06 dB PSNR improvement, confirming the method’s ability to mitigate local misalignments. Unlike GaussReg, which relies on scene-center proximity for Gaussian selection, the proposed approach enforces spatial consistency through structural priors, reducing blending artifacts while preserving fine-scale features.

On the Coord dataset, the structured skeleton prior improves Gaussian selection in overlapping regions, achieving a 0.016 SSIM increase and a 3.1\% LPIPS reduction. In contrast, GaussReg discards critical Gaussians in high-overlap areas, causing blurred textures and inconsistent depth cues. These results validate that skeleton-aware fusion effectively balances global consistency and local detail preservation, mitigating structural distortions and radiometric inconsistencies.

Fig.~\ref{fig:fusion_comparison} showcases three sub-maps fusion examples, including our Coord scene and two scenes from the ScanNet-GSReg dataset. Each scene is fused with either GaussReg or the proposed approach. In GaussReg, which relies on the scene-center distance for merging decisions, critical Gaussians located away from the center or in high-overlap regions can be incorrectly discarded, creating gaps in the final reconstruction. Additionally, the absence of structural guidance in GaussReg often leads to local misalignments and blurred details, as spatially overlapping Gaussians from different viewpoints may not be properly integrated. By contrast, the skeleton-based fusion leverages the global skeleton to preserve useful Gaussians, promoting structural coherence and sharper rendering in overlapping areas. This difference is most evident near architectural edges and high-curvature regions, where local detail is essential for visual fidelity.

\section{CONCLUSIONS}
We present an automated 3D-GS registration and fusion framework to address viewpoint disparity, structural misalignment, and feature inconsistency in ground-aerial reconstruction. A structure-aware skeleton alignment ensures robust initialization, while GA-KPConv enhances feature extraction using Mahalanobis distance-based neighborhood selection and ellipsoid-aware convolution. A multi-factor Gaussian fusion strategy optimally selects Gaussians by balancing structural adherence, detail preservation, and spatial consistency, producing a unified high-fidelity reconstruction. Extensive experiments confirm improvements in registration accuracy, structural coherence, and reconstruction fidelity.

Despite its effectiveness, the framework has limitations. Feature extraction primarily focuses on geometry, lacking semantic awareness. Future work will explore adaptive clustering and hybrid geometric-semantic fusion to enhance robustness in complex scenes.

\bibliographystyle{unsrt}
\bibliography{ref}

\end{document}